\documentclass{iucrjournals}

\usepackage{amsmath}

\title{Domain Knowledge Guided Bayesian Optimization For Autonomous Alignment Of Complex Scientific Instruments}

\author{Aashwin Mishra\IUCrCemaillink{aashwin@slac.stanford.edu}}
\author{Matt Seaberg\IUCrCemaillink{seaberg@slac.stanford.edu}}
\author{Ryan Roussel\IUCrCemaillink{rroussel@slac.stanford.edu}}
\author{Daniel Ratner\IUCrCemaillink{dratner@slac.stanford.edu}}
\author{Apurva Mehta\IUCrCemaillink{mehta@slac.stanford.edu}}
\affil{SLAC National Laboratory, 2575 Sand Hill Rd, Menlo Park, CA 94025, USA}

\begin{document} 
\nolinenumbers
\maketitle 

\begin{abstract}
 Bayesian Optimization (BO) is a powerful tool for optimizing complex non-linear systems. However, its performance degrades in high-dimensional problems with tightly coupled parameters and highly asymmetric objective landscapes, where rewards are sparse. In such needle-in-a-haystack scenarios, even advanced methods like trust-region BO (TurBO) often lead to unsatisfactory results. We propose a domain knowledge guided Bayesian Optimization approach, which leverages physical insight to fundamentally simplify the search problem by transforming coordinates to decouple input features and align the active subspaces with the primary search axes. We demonstrate this approach's efficacy on a challenging 12-dimensional, 6-crystal Split-and-Delay optical system, where conventional approaches, including standard BO, TuRBO and multi-objective BO, consistently led to unsatisfactory results. When combined with an reverse annealing exploration strategy, this approach reliably converges to the global optimum. The coordinate transformation itself is the key to this success, significantly accelerating the search by aligning input co-ordinate axes with the problem's active subspaces. As increasingly complex scientific instruments, from large telescopes to new spectrometers at X-ray Free Electron Lasers are deployed, the demand for robust high-dimensional optimization grows. Our results demonstrate a generalizable paradigm: leveraging physical insight to transform high-dimensional, coupled optimization problems into simpler representations can enable rapid and robust automated tuning for consistent high performance while still retaining current optimization algorithms.

\end{abstract}

\keywords{Optimization, Bayesian Optimization, Domain Informed Machine Learning, Physics Informed Machine Learning, Beamline Tuning}

\section{Introduction}
The Fourth Paradigm of Science \cite{agrawal2016perspective, tolle2011fourth, hey2009fourth} is data-intensive discovery and drives development of a new generation of large-scale scientific facilities. Instruments like the Large Hadron Collider, the James Webb Space Telescope and next-generation of large of light sources are pushing the boundaries of scientific knowledge. However, the operational complexity of these facilities presents a substantial bottleneck. These are not static devices but high-dimensional, dynamic systems where hundreds of coupled parameters must be continuously optimized to maintain optimal scientific throughput. To this end, relying on manual tuning using human experts is becoming increasingly untenable. Such manual tuning is slow, lacks reproducibility and consumes valuable time that could be used for discovery. This operational gap highlights a critical need for intelligent agents that can autonomously navigate these complex parameter spaces, creating a new frontier for machine learning in the form of self-driving laboratories. Developing the robust, sample-efficient algorithms required to control these complex instruments is therefore a cornerstone challenge for accelerating scientific progress in the 21st century.

A central task in the autonomous control of these instruments is high dimensional black box optimization. Frequently, the objective functions exhibit a needle in a haystack characteristic \cite{adricioaei1996finding, kim2020machine}. Here the region of optimal performance is exceptionally sparse and provides near-zero reward across the vast majority of the parameter space. This difficulty is severely compounded by the strong physical interdependence between the control parameters, a phenomenon that is ubiquitous in scientific domains. For instance, correcting the beam path in a particle accelerator requires co-optimizing hundreds of steering magnets whose effects are coupled through the machine's lattice \cite{bettoni2025machine}. Similarly, in adaptive optics, the overlapping influence functions of a deformable mirror's actuators necessitate a coordinated, non-local control signal to correct wavefront aberrations\cite{haber2021general}. In fusion research, stabilizing a plasma inside a tokamak depends on carefully balancing and simultaneously adjusting numerous magnetic coils together\cite{degrave2022magnetic}. In the language of machine learning, this physical coupling means the objective function’s active subspaces \cite{constantine2015active} are not aligned with the native control axes, forcing the solution to lie on a narrow, non-axis-aligned ridge. Solving this compound problem: locating a sparse needle that lies along a narrow, misaligned ridge presents a fundamental challenge that pushes even state of the art black box optimization algorithms to their limits. This challenge grows exponentially with the inclusion of each new parameter.

To highlight these challenges, we select a pertinent real-world benchmark problem involving the Hard X-ray Split-and-Delay (HXRSND) system at the Linac Coherent Light Source (LCLS) X-ray laser \cite{Schoenlein2016,Zhu2017}. The HXRSND uses a cascade of six crystals to split a single X-ray pulse into two, introduce a precise delay to one path and then perfectly recombine the two paths in both space and angle. This system serves as an ideal testbed as it simultaneously manifests all the aforementioned optimization challenges. Its control space is high dimensional, explicitly 12-dimensional, comprising the pitch and roll angles for each of the six crystals. The physical requirement for Dynamical Scattering Bragg diffraction at each crystal, represented by a sharp table-top Darwin acceptance along the scattering direction, coupled with a relatively shallow optimal region in the orthogonal direction, creates the archetypal needle in a haystack objective (see Fig 2). Even maintaining diffraction condition for each crystal, without which no X-ray beam will propagate beyond the crystal, demands nano-radian precision within a search space that is thousands of times larger. Finally, the sequential nature of the optics introduces extreme coupling where an angular adjustment to any single crystal alters the beam's trajectory, effectively misaligning the entire downstream system and forcing the solution onto the narrow, winding ridge of global optimality. Successfully aligning the HXRSND is therefore not just an operational necessity, but a rigorous validation for any optimization algorithm claiming to solve complex, coupled problems.

For such optimization problems, classical local search methods like the Nelder-Mead (Simplex) algorithm \cite{nelder1965simplex} falter, making slow, zigzagging progress as they attempt to trace the narrow optimal ridge. Given its sample efficiency, Bayesian Optimization (BO) \cite{shahriari2015taking, frazier2018bayesian, garnett2023bayesian} is a natural candidate for this task, yet it proves insufficient. A classical BO procedure, which models the objective with a Gaussian Process, struggles because standard kernels are inherently axis-aligned. While such a model can eventually learn the strong off-diagonal correlations defining the optimal ridge, it requires a prohibitive number of samples, defeating the purpose of sample-efficient optimization. We show that even more advanced methods designed for such challenges, like the state of the art Trust Region BO (TurBO) \cite{eriksson2019scalable}, fail to converge to the global optimum within the allocated sample budget. TurBO’s strategy of exploring within axis-aligned, hyper-rectangular trust regions is fundamentally mismatched with the diagonal geometry of such problems with coupling amongst input features. The algorithm wastes precious function evaluations exploring the corners of this hyper-rectangle, which lie far from the optimal path, leading to a failure to find improvement and a premature, unproductive shrinking of the trust region. The core issue is therefore not a flaw in the search strategy itself, but a fundamental mismatch between the problem's intrinsic physical geometry and the implicit assumptions of the algorithm.

In this work, we outline a domain knowledge guided approach to Bayesian Optimization that directly addresses this fundamental mismatch between the optimization algorithm and the optimization problem. Instead of treating the system as an intractable black box, our method leverages domain physics knowledge to apply coordinate transformations that decouple the dominant physical effects, aligning the problem's narrow active subspaces with the optimizer's search axes. We pair this transformation with a reverse annealing schedule that enforces the sustained exploration needed to locate the sparse global optimum. Through a rigorous case study on the 12-dimensional HXRSND system, we demonstrate that this domain knowledge guided approach consistently converges to the desired solution, a task where traditional and even state of the art trust region BO methods fail. Our primary contributions are therefore: 
\begin{enumerate}
    \item a novel, two-pronged approach combining a physics-derived coordinate transformation with a reverse annealing schedule to solve coupled, needle in a haystack problems.
    \item a rigorous demonstration on a complex, real-world benchmark showing that this domain knowledge guided BO succeeds where other SOTA methods are unsatisfactory.
    \item a generalizable blueprint for creating intelligent optimization agents for other complex scientific and physical systems defined by strong parameter inter-dependencies.
\end{enumerate}

Finally, while this work demonstrates efficacy on a specific X-ray optical system, the underlying principle of decoupling physical symmetries is universal. We distill this approach into a generalized ``recipe" for applying domain-guided coordinate transformations to other coupled physics systems, which is provided in Appendix A.

\section{Methods \& Details}

\subsection{Details of the HXRSND}
\begin{figure}[h!]
\centering
\includegraphics[width=\textwidth]{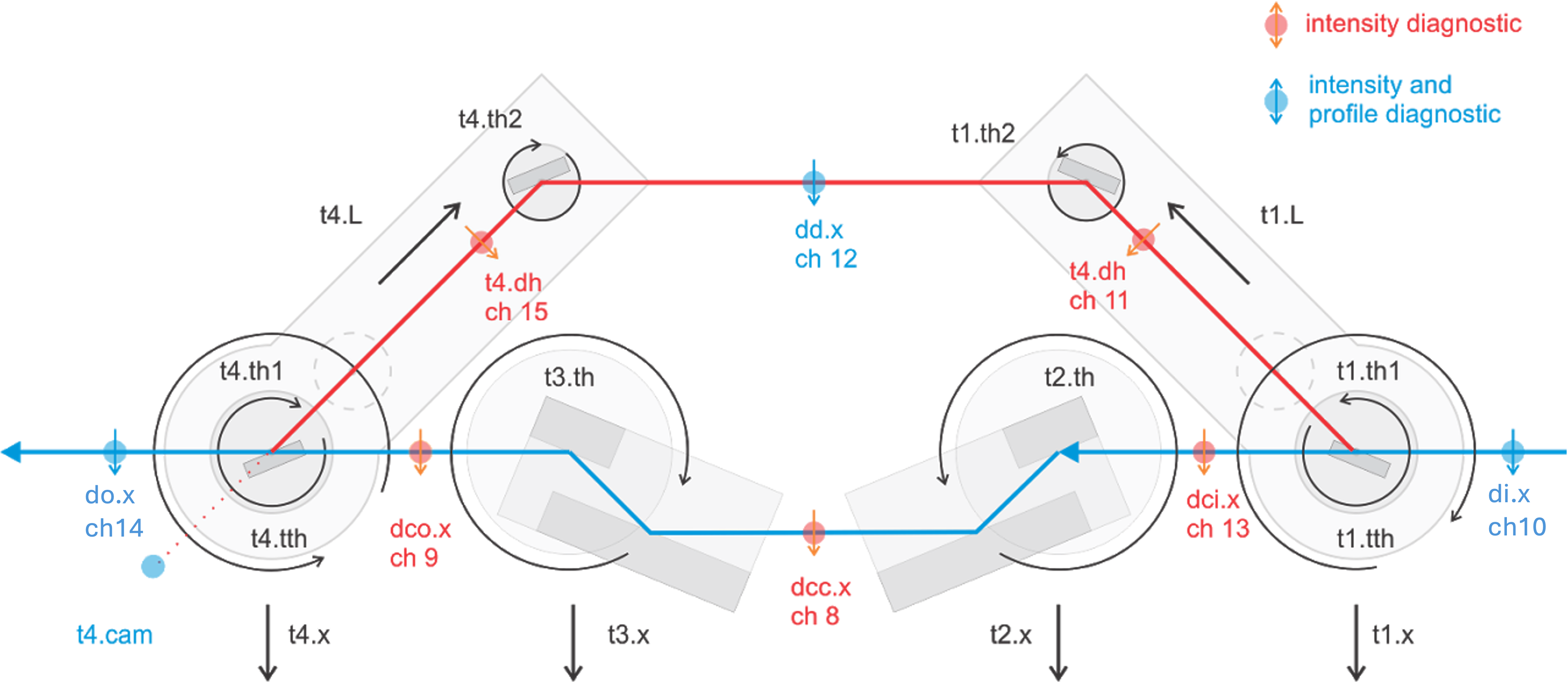}
\caption[]{Schematic of the HXRSND with the CC (Channel Cut) branch in blue and the delay branch in red. The x-ray beam propagates from right to left. The arrows indicate the motorized degrees of the system. In addition, each crystal along the delay branch has ``chi'' adjustment which corresponds to rotation about the tangential vector of the crystal surface (not shown). The red and blue dots correspond to locations of beam diagnostics, as noted in the legend on the top right of the figure \cite{Zhu2017}.}
\label{fig:fig1}
\end{figure}

Split and Delay systems are common and critical components of modern beamlines particularly at X-ray Free-Electron Laser (XFEL) and Extreme Ultraviolet (XUV) facilities \cite{rysov2019compact, roling2012design}. They serve as the ``optical counterparts" to traditional laser delay lines, enabling ultrafast science by splitting a single pulse into two phase coherent or mutually coherent copies to perform time domain measurements. 
As outlined in Figure \ref{fig:fig1}, the HXRSND consists of two branches: the minimally adjustable “channel-cut” (CC) branch outlined in blue, and the “delay” branch with twelve degrees of freedom outlined in red. This delay range lies between 5 to 500 ps. With the advent of LCLS-II-HE upgrades \cite{LCLSIIHE, mishra2025start}, this HXRSND optics will be critical for ultrafast studies of complex materials. We, therefore, must ensure that operational inefficiencies and system limitations do not become a bottleneck for these experiments. The alignment of the HXRSND requires a spatial overlap between the two branches at the sample with very high precision, along with optimized intensity at the output.  As an illustration, for X-ray Photon Correlation Spectroscopy (XPCS) where the HXRSND can provide twin pulses with custom delays, both the branches must be aligned to the same photon energy to within ~0.1 eV, with almost perfect overlap and matched intensities from the two branches. Currently, the HXRSND is aligned sequentially (from the right to the left in Figure \ref{fig:fig1}), by an experienced beamline scientist, using intermediate sensors, as even expert operators cannot effectively co-optimize more than two or three dimensions simultaneously. It takes an experienced operator up to 4 hours for alignment, and the final setting is often far from optimal. 

\begin{figure}[h!]
\centering
\includegraphics[clip, trim=5cm 6cm 9.5cm 5cm, width=\textwidth]{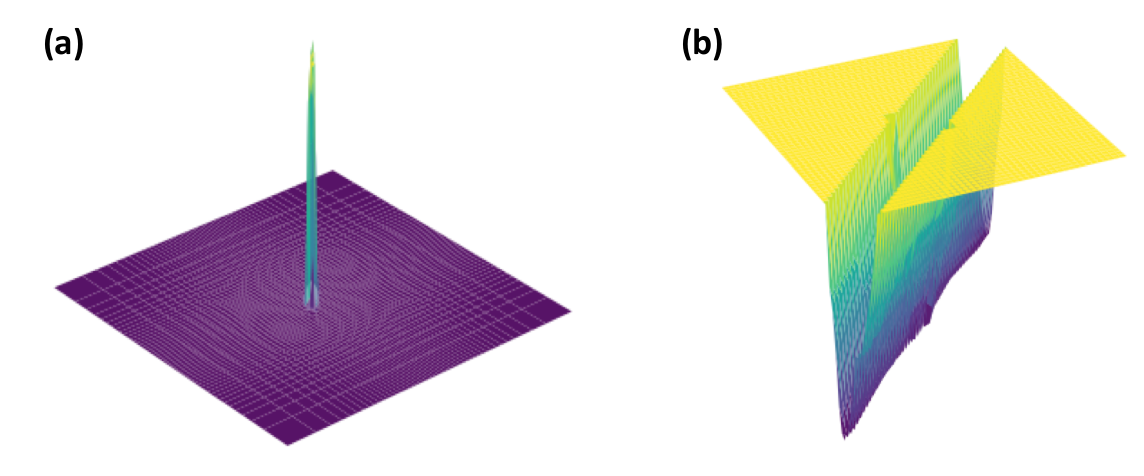}
\caption[]{Schematic outlining the key challenges in the optimization of the HXRSND, (a) Contour surfaces of the beam intensity in a 2D subsection of the 8D space exhibiting the sparsity of the optimum and the lack of informative gradients, (b) Contour surfaces of the beam position error exhibiting the interdependence between input features. For both (a) and (b), the inputs correspond to the t1.th1 and t1.th2 settings varying over $\pm 100$ microrads (see Fig.~\ref{fig:fig1}).}
\label{fig:fig2}
\end{figure}
To achieve precise alignment for the HXRSND the spatial overlap of both branches at the sample must be manifest along with optimized output intensity. Consequently, minimizing beam position error between the branches is critical for ensuring overlap. Simultaneous adjustments are also required to maximize beam intensity. The distinct optimization landscapes of these two optimization objectives (beam intensity and beam position error) can be traced directly to their underlying physics. The intensity objective exhibits a classic needle in a haystack profile because the final throughput is a multiplicative function of each crystal's reflectivity. Governed by the narrow Bragg condition, each crystal acts as a ``gate" that is open only within a micro-radian scale window, resulting in a sparse reward landscape where the signal is zero unless all crystals are simultaneously aligned as illustrated in Figure \ref{fig:fig2} (a). Conversely, the beam position error is defined by strong interdependence arising from geometric optics. This objective is a cumulative function of the entire beam path, where adjusting one crystal perturbs the alignment of all subsequent optics, creating a coupled problem where the solution lies along a narrow, non-axis-aligned ridge in the parameter space, as illustrated in Figure \ref{fig:fig2} (b).

\subsection{Outline of domain knowledge guided Bayesian Optimization approach}

A central obstacle in optimizing multi-element optical systems is the strong interdependence between control parameters. As established with the HXRSND, an adjustment to an upstream component (for instance changing the pitch angle of the first crystal) propagates downstream and alters both the position and angle of the X-ray beam incident on all subsequent optics. This coupling restricts the manual optimization procedure to have to function sequentially, element by element along the knobs. For a human operator, this manifests as a non-intuitive process of walking multiple knobs simultaneously, where optimizing one parameter actively degrades the performance of others. This is the same failure mode exhibited by naive optimization algorithms like coordinate descent or greedy hill-climbing that make slow, zigzagging progress across the optimal path. Even more robust local search methods like the Nelder-Mead algorithm struggle to efficiently trace such features. In mathematical terms, this physical interdependence means the objective function's active subspaces \cite{constantine2014active} (the low-dimensional manifolds where the function exhibits significant variation) are not aligned with the input coordinate axes. Instead, the region of low beam position error forms a narrow, non axis aligned ridge within the high dimensional parameter space, making it exceptionally difficult to trace.

This absence of axis alignment in the structure poses a severe challenge even for sophisticated black box optimizers. A classical Bayesian Optimization procedure, which models the objective using a Gaussian Process (GP)\cite{rasmussen2006gaussian}, struggles because standard kernels (e.g., a RBF kernel) are inherently axis-aligned. Using Automatic Relevance Determination (ARD) enables different length scales along each axis. However, such ARD can only account for axis scaling but not misalignments between the input axes and the active subspaces. While a GP can eventually learn off-diagonal correlations, it requires a large number of samples to do so which defeats the purpose of sample efficient optimization. More advanced trust region methods like TurBO, which are specifically designed to handle challenging high-dimensional problems, also exhibit limited efficacy. TurBO achieves efficiency by maintaining local models within axis-aligned hyper-rectangular trust regions. However, when the optimal path is a narrow ridge oriented diagonally to these axes, the rectangular trust region is a fundamentally inefficient geometry. The algorithm wastes precious samples exploring the corners of the rectangle, which lie far from the optimal ridge, leading to a failure to find improvement and a premature, unproductive shrinking of the trust region. The core issue is therefore not a failure of the BO search strategy itself but a fundamental mismatch between the problem's intrinsic geometry and the algorithm's implicit, axis-aligned assumptions. This motivates our primary contribution: \textit{instead of forcing the algorithm to learn this complex structure, we use domain knowledge to transform the problem itself into one that adheres to the assumptions of BO algorithms}. This makes the problem easier to solve.

While the domain knowledge based transformation effectively accounts for the interdependence dominating the beam position error, the full optimization task for the HXRSND remains multi-objective which is to minimize this error while simultaneously also maximizing beam intensity. We approach this via a scalarized objective, which is well-posed because the manifold of minimal beam position error (where the two beam paths spatially overlap) is a necessary precondition that contains the much sparser region of maximum intensity. However this creates a refined needle in a smaller haystack problem. Even with the smaller search space simplified by the transformation, a standard Bayesian Optimization algorithm, on finding the broad basin of attraction corresponding to low beam position error, will naturally reduce its exploration to exploit this promising region. It is therefore prone to premature convergence on a suboptimal solution with perfect beam overlap but negligible intensity, having never discovered the narrow intensity peak hidden within.

To overcome this tendency for premature convergence, we introduce a second component to the domain knowledge guided BO approach, a scheduled exploration strategy of reverse annealing. We employ the Upper Confidence Bound (UCB) as our acquisition function which guides the search by balancing the predicted mean of the objective (exploitation) with the model's uncertainty (exploration) moderated by a hyperparameter $\beta$. In a conventional annealing schedule, $\beta$ is decreased over time to favor exploitation as the model's certainty grows with increasing samples to characterize the input-output relationship. With reverse annealing, we propose the opposite approach. By monotonically increasing $\beta$ with each iteration, our reverse annealing schedule forces the optimizer to also prioritize regions of high uncertainty. This strategy actively counteracts the algorithm's natural inclination to over exploit the initial good solutions that it finds. By compelling continued exploration even late in the optimization process, we significantly increase the probability of discovering the elusive needle of high intensity that resides within the broader, but ultimately suboptimal region of low beam position error.

It is important to distinguish our domain knowledge guided transformation approach from other methods that incorporate prior knowledge into Bayesian Optimization, most notably the use of learnt kernels \cite{duris2020bayesian, hanuka2021physics, edelen2019machine}. The key distinction lies in \textit{how} prior knowledge is used: our method simplifies the problem itself, whereas the alternative builds a more complex model for the problem. In the learnt kernel approach, a custom Gaussian Process kernel is pre-trained on an available dataset, which may be sourced from archival experimental data or physics simulations. The aim is for this complex kernel to learn and implicitly encode the underlying covariance structure of the objective function, including the non-axis-aligned interdependencies between parameters. Often, this pre-trained kernel is then ``frozen" (that is, the parameters of the kernel are fixed and not changed during the optimization procedure.) and used within a standard Bayesian Optimization loop, providing a data-driven prior that is intended to guide the search more efficiently than a generic kernel. The domain knowledge guided approach discussed and outlined in this investigation, differs fundamentally in both philosophy and execution, offering several distinct advantages. First and most critically, our method is not dependent on pre-existing datasets. Rather than using data to build a complex model of a complex space, we use physics principles and domain knowledge to apply coordinate transformations that simplify the space itself. This is an advantage in scientific domains where historical data may be sparse, expensive to generate or not representative of current machine conditions. Second, the physics based transforms are amenable to incorporation with any Bayesian Optimization algorithm (and even with general optimization approaches). Conversely, integrating structured, non-stationary kernels into trust-region frameworks like TurBO is mathematically non-trivial and computationally prohibitive. Finally, our method is more computationally efficient and interpretable. It allows the BO algorithm to operate with a simple, standard, axis-aligned kernel within the transformed space. This avoids the computational overhead of a dense, custom covariance matrix. The transformation itself is physically meaningful, making the optimization process transparent and easier to debug. On the other hand, a learnt kernel may often function as a less-interpretable abstraction.

\subsection{Application to the HXRSND System}
The optimization of the HXRSND involves tuning multiple rotational degrees of freedom for each of the six crystals, primarily the pitch (theta) and roll (chi) angles. The theta motion governs the Bragg condition and is responsible for the needle in a haystack (sparse) character of the intensity objective, requiring nano-radian precision. The chi motion, which controls the planarity of reflections, is a primary source of the strong interdependence between crystals. An error in chi on any one of the crystals breaks the alignment for all subsequent crystals, making independent tuning impossible and creating the non axis aligned optimization landscape that challenges conventional BO algorithms.

The specific structure of the interdependence in the beam position error objective can be predicted directly from the HXRSND's design. The instrument splits the beam into a shorter Channel Cut (CC) path and a longer, variable Delay Path, with the goal of recombining them perfectly. The objective is therefore a differential measurement: the spatial error between the outputs of these two distinct optical branches. We can consider two corresponding knobs, $k_{ref}$ and $k_{delay}$, controlling elements on the beam path. The system is designed such that a perturbation to $k_{ref}$ induces a positional change in its beam that is functionally similar to the change induced by the same perturbation to $k_{delay}$. The condition for constant position error (the definition of a contour line) is therefore when the perturbations to each knob are approximately equal. In the 2D parameter space, this relationship describes a line at a 45-degree angle to the native axes. This allows us to redefine the problem in terms of a highly sensitive differential mode ($v_{diff} = k_{delay} - k_{ref}$) which controls the position error and a largely insensitive common mode ($v_{common} = k_{delay} + k_{ref}$) which steers both beams together. Physics thus predicts that the optimization landscape for position error should be a narrow canyon aligned diagonally to the control axes.

The pairings between different control knobs can be illustrated from Figure \ref{fig:fig1}. First, we discuss the delay branch, outlined in red in the figure. The first paired knobs are labeled $t1.th1$ and $t4.th1$, and control the pitch of the first and the fourth crystals, respectively. Physically, these two crystals control the overall input and output beam angle of the chicane. To ensure that the exiting beam is parallel to the incoming beam, their pitch angles are strongly anti-correlated. Any changes on $t1.th1$ must be matched by a corresponding change on $t4.th1$ to maintain the final beam direction. Similarly, another set of paired knobs are labeled $t1.th2$ and $t4.th2$ in Figure \ref{fig:fig1}, controlling the pitch of the second and the third crystals, respectively. These two crystals control the path within the chicane. Their pitch angles are also strongly anti-correlated to keep the beam path parallel through the middle section of the chicane. Similar pairings of the knobs exist for alignments in the vertical direction (or out of the plane shown in Figure \ref{fig:fig1}). The paired knobs $t1.chi1$ and $t4.chi1$, need to be balanced to prevent the beam from walking off vertically as it enters and exits the chicane. This is a primary source of Beam Position Error. Similarly, the knobs $t1.chi2$ and $t4.chi2$, need to be balanced against each other to ensure that the beam does not acquire a vertical ``kink" in the middle of the delay path. Similar pairings are evident in the channel cut branch (outlined in blue in the figure), where the ($t2.th$, $t3.th$) and the ($t2.chi$, $t3.chi$) knobs need to be adjusted relative to each other so as to ensure a correct beam path for the channel cut beam branch.

\begin{figure}[h!]
\centering
\includegraphics[clip, trim=1.5cm 4cm 1cm 4cm, width=\textwidth]{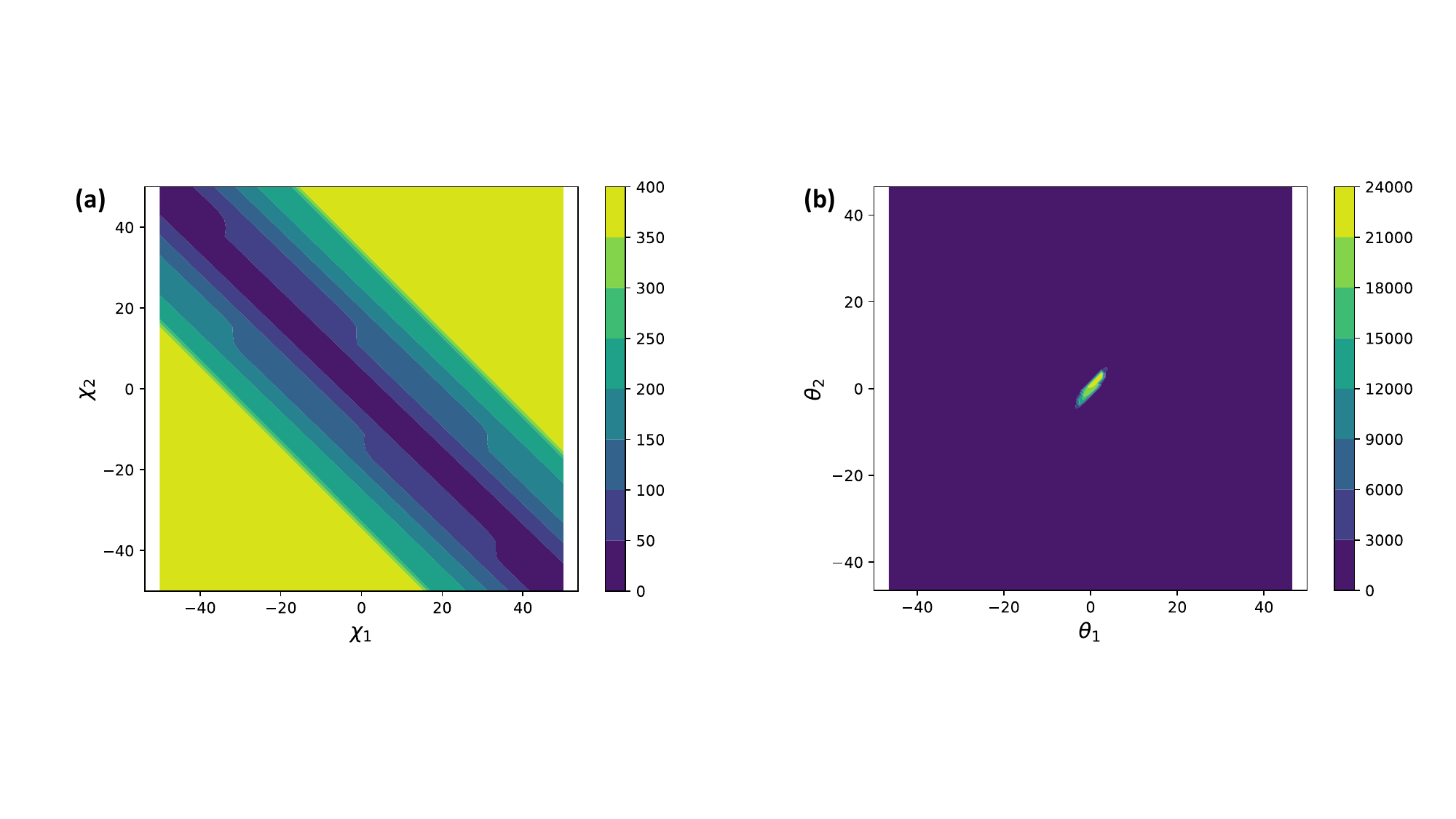}
\caption[]{Schematic outlining (a) the interdependence between inputs in the Beam Position Error objective in $\mu$m (microns) (b) the sparsity of the Beam Intensity objective over the input space in arbitrary units (a.u.). For both (a) and (b), the inputs perturbations are marked in units of microrads.}
\label{fig:fig3}
\end{figure}

To validate this physical intuition using numerical simulations, we evaluated the objective functions across two-dimensional slices of the 12-dimensional parameter space. Figure \ref{fig:fig3} presents these results. As predicted, the contour plot for the beam position error (Figure \ref{fig:fig3} (a)), shown as a function of two corresponding chi knobs, reveals distinct bands of low error oriented precisely at 45 degrees to the input axes. This provides empirical evidence that the optimization problem is dominated by a non axis aligned active subspace. In contrast, Figure \ref{fig:fig3} (b) shows the intensity objective as a function of two theta knobs. The contour surface for the beam intensity is characterized by a single, highly localized peak of high intensity, visually confirming the sparse needle in a haystack nature of this second objective. The diverse yet coupled complexity in  this problem (a diagonal ridge for positional alignment that also containing a sparse peak for intensity) presents a compound challenge that conventional optimizers may be ill equipped to handle. This highlights the need for a multi-pronged approach for optimization. We outline such an approach in the two part strategy detailed in the following sections.

The observation of a 45-degree interdependence within each pair of corresponding knobs motivates a coordinate transformation that rotates the problem into a more natural, physically meaningful basis. We move from the native 12D space of individual knob settings, $(k_{CC.1}, k_{delay.1}, ..., k_{CC.6}, k_{delay.6})$, to a new 12D space defined by the common and differential direction of each pair. For each $i^{th}$ pair, this transformation is a 2D rotation:
\begin{equation}
\mathbf{v}_i
=
\begin{bmatrix}
v_{\text{diff}, i} \\
v_{\text{common}, i}
\end{bmatrix}
=
\frac{1}{\sqrt{2}}
\begin{bmatrix}
1 & -1 \\
1 & 1
\end{bmatrix}
\begin{bmatrix}
k_{\text{delay}, i} \\
k_{\text{ref}, i}
\end{bmatrix}
=
\mathbf{R} \mathbf{k}_i
\end{equation}
This 2D rotation is applied independently to each of the six pairs, constituting a full 12x12 block-diagonal linear transformation. The resulting 12D vector v = $(v_{diff,1}, v_{common,1}, ..., v_{diff,6}, v_{common,6})$ represents our new, transformed input space.

This transformation achieves the exact goal required for efficient optimization. In this space, the beam position error objective is almost exclusively dependent on the six differential mode axes $(v_{diff,i})$, while being largely insensitive to changes along the six common mode axes $(v_{common,i})$. By aligning the active subspace of the position error objective with the coordinate axes, we enable an axis-aligned optimizer like TurBO to function with maximum efficiency.

The transformation defined above is derived specifically for the geometry of the HXRSND. However, the logic used to identify these modes is transferable to any system with coupled inputs. Readers interested in deriving similar transformations for their own physics or engineering systems may refer to Appendix A, where we outline a step-by-step recipe for this domain guided approach.

Having aligned the active subspace of the position error with the coordinate axes, the remaining challenge is to locate the sparse optima of beam intensity within this smaller and tractable region. A standard Bayesian Optimization approach using an Upper Confidence Bound (UCB) acquisition function,
\begin{equation}
a(\mathbf{x}) = \mu(\mathbf{x}) + \sqrt{\beta} \sigma(\mathbf{x}),
\label{eq:ucb}
\end{equation}
is likely to fail here. The algorithm will quickly identify the broad basin corresponding to low position error, a region of high predicted mean $\mu(\mathbf{x})$. With a conventional fixed or decreasing $\beta$ schedule, the optimizer will prematurely converge, ceasing exploration to exploit this large suboptimal region and thereby never discovering the narrow intensity peak hidden within.
To counteract this tendency, we employ a reverse annealing schedule where the exploration-exploitation hyperparameter $\beta$ is monotonically increased with each iteration. This strategy forces the algorithm to become progressively more exploratory over time, giving increasing weight to the model uncertainty $\sigma(\mathbf{x})$. Even after converging on the region of low position error, the optimizer is thus compelled to continue sampling locations of high uncertainty within that region which is precisely where the undiscovered intensity peak is expected to lie. After empirical evaluation, we found a simple linear schedule,
\begin{equation}
\beta(t) = \beta_0 + c \cdot t,
\label{eq:rev_anneal}
\end{equation}
where $t$ is the iteration number, provides the sustained exploratory pressure required to consistently overcome the local optimum and locate the true, high-intensity global optimum. For this work, we used parameter values of $\beta_0 = 1.0$ and $c = 0.01$.

In the next section, we compare and contrast the results from the domain knowledge guided Bayesian Optimization approach against results from using standard Bayesian Optimization (BO), Trust Region Based Bayesian Optimization (TurBO) and Multi-Objective Bayesian Optimization (MOBO). To enable a direct comparison using single-objective optimizers (against standard BO and TurBO), we formulated a scalarized objective that combines the two competing goals: minimizing the beam position error and maximizing the beam intensity. The first critical step is normalization, required to bring the two objectives, which have disparate units and scales, onto a common footing. We scaled both the position error, $E(\mathbf{x})$, and the intensity, $I(\mathbf{x})$, to the range $[0, 1]$ using min-max normalization based on empirically determined bounds:
\begin{align}
E_{\text{scaled}}(\mathbf{x}) &= \frac{E(\mathbf{x}) - E_{\text{min}}}{E_{\text{max}} - E_{\text{min}}} \\
I_{\text{scaled}}(\mathbf{x}) &= \frac{I(\mathbf{x}) - I_{\text{min}}}{I_{\text{max}} - I_{\text{min}}}
\end{align}
This ensures that neither objective dominates the scalarized function due to its magnitude alone.
With the objectives normalized, we constructed a single minimization objective, $f(\mathbf{x})$, by subtracting the scaled intensity from the scaled position error:
\begin{equation}
f(\mathbf{x}) = E_{\text{scaled}}(\mathbf{x}) - I_{\text{scaled}}(\mathbf{x})
\label{eq:scalar_obj}
\end{equation}
This formulation naturally rewards solutions with low error and high intensity. The optimal value for this function is $-1$, corresponding to zero position error ($E_{\text{scaled}}=0$) and maximum intensity ($I_{\text{scaled}}=1$). While we experimented with different weighting schemes, we found this simple, equally weighted form to be the most effective. The normalization step already provides the necessary balancing, and this form avoids introducing arbitrary weighting parameters that could bias the search. All subsequent single-objective optimization results are based on the minimization of $f(\mathbf{x})$ as defined in Eq.~\ref{eq:scalar_obj}.

For the Multi-Objective Bayesian Optimization (MOBO) baseline, the two objectives were handled directly without scalarization, allowing for a more complete exploration of the trade-offs between them. The goal of MOBO is not to find a single optimal point, but to approximate the Pareto front: the set of solutions where one objective cannot be improved without degrading the other. To guide this search, we employed the widely-used Expected Hypervolume Improvement (EHVI) acquisition function. EHVI selects the next evaluation point that is expected to yield the largest increase in the hypervolume dominated by the current estimated Pareto front. The optimization was posed as a dual-minimization problem by minimizing the beam position error, $E(\mathbf{x})$, and simultaneously minimizing the negative of the intensity, $-I(\mathbf{x})$.


\section{Results \& Discussion}
\begin{figure}[h!]
\centering
\includegraphics[width=\textwidth,trim={6.25cm 2.05cm 6.25cm 2cm},clip]{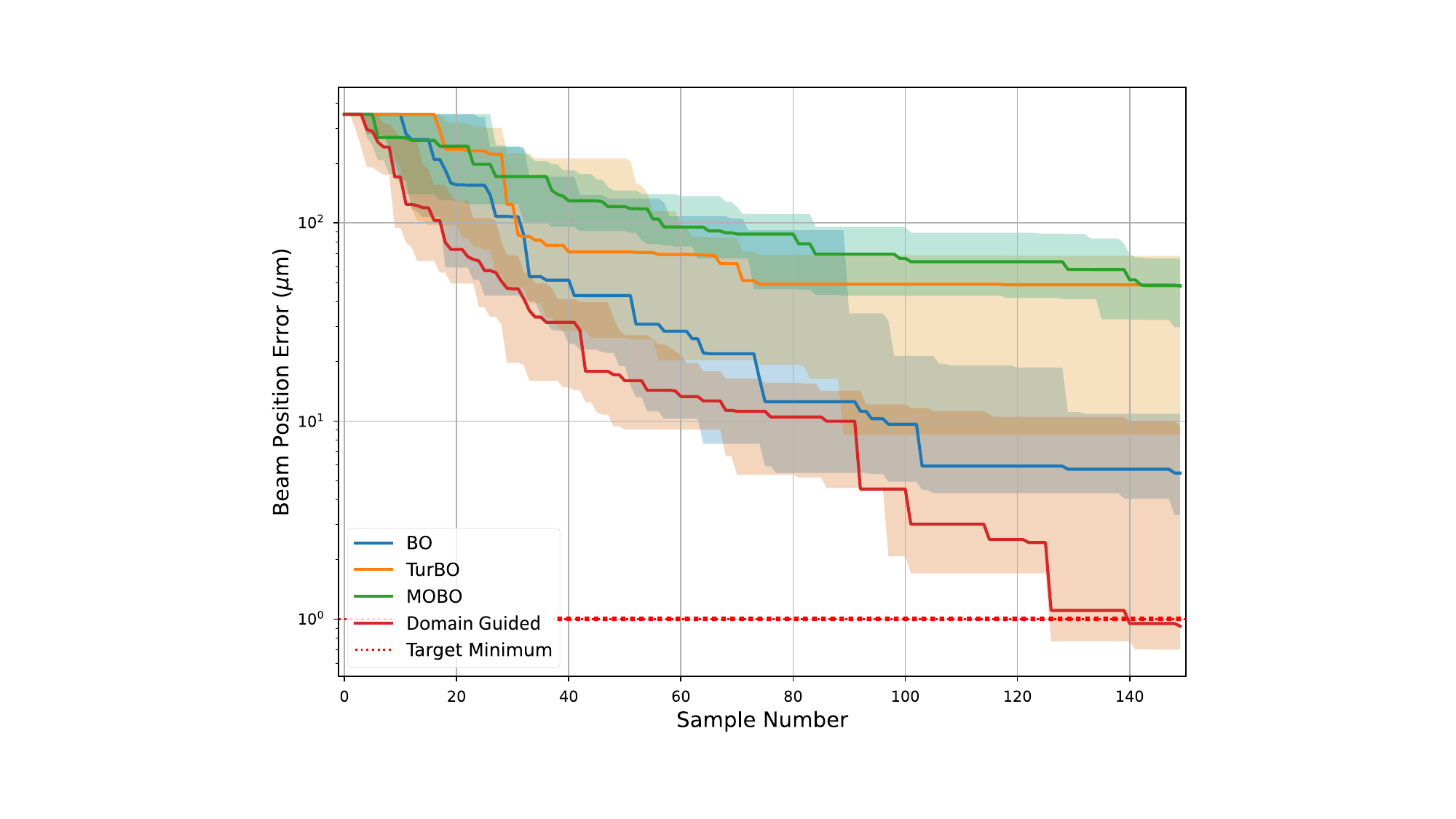}
\caption[]{Comparison between the performance of domain knowledge guided Bayesian Optimization (labeled as Domain Guided), against standard Bayesian Optimization (BO), Trust Region based Bayesian Optimization (TurBO) and Multi-Objective Bayesian Optimization (MOBO) for the Beam Position Error. For each algorithm, we report the median of the running minimum with the solid line, over $25$ experiments. The shaded zones outline the $25^{th}$ and $75^{th}$ percentiles.}
\label{fig:bpe_comparison}
\end{figure}

In this section, we compare and contrast the proposed domain knowledge guided Bayesian Optimization approach's results against vanilla Bayesian Optimization (BO), Trust Region based Bayesian Optimization (TurBO) and Multi-Objective Bayesian Optimization (MOBO) when applied to the optimization of the HXRSND. All single-objective baselines (standard BO, TurBO and the domian knowledge guided BO) utilized the same scalarized objective function defined in Eq. 6 to ensure a fair comparison. To account for the stochastic nature of the optimization algorithms, we performed $25$ independent trials for each method (the proposed domain knowledge guided Bayesian Optimization approach, TurBO, standard BO, and MOBO), each starting from a different set of initial random samples. The total length of each independent run was set to $150$ samples. Each run started with $4$ initial random samples, and these were identical across all the four algorithms. The performance metrics reported in this section are aggregated over these 25 trials.

\begin{figure}[h!]
\centering
\includegraphics[width=\textwidth,trim={6.25cm 2.05cm 6.25cm 2cm},clip]{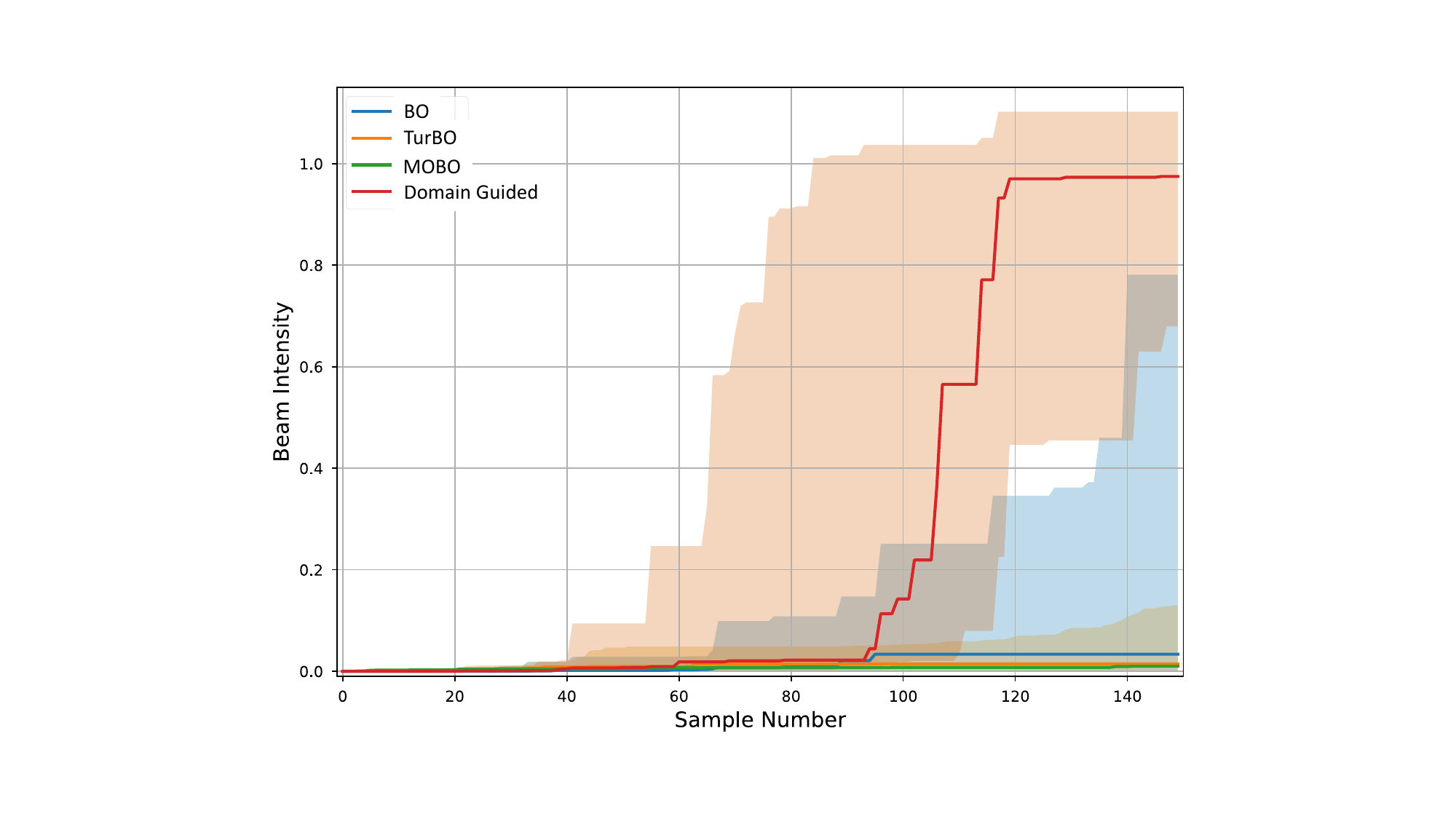}
\caption[]{Comparison between the performance of domain knowledge guided Bayesian Optimization (labeled as Domain Guided), against standard Bayesian Optimization (BO), Trust Region based Bayesian Optimization (TurBO) and Multi-Objective Bayesian Optimization (MOBO) for the Beam Intensity. For each algorithm, we report the median of the running maximum with the solid line, over $25$ experiments. The shaded zones outline the $25^{th}$ and $75^{th}$ percentiles.}
\label{fig:intensity_comparison}
\end{figure}

In Figure \ref{fig:bpe_comparison}, we report the results over these sets of trials for each algorithm for the minimization of the Beam Position Error. Based on the sizes of the CC and delay branches, we need a threshold beam position error to ensure overlap, which is labeled as the ``Target Minimum" in the figure. We observe that the TurBO algorithm exhibits less than satisfactory results, which can be explained from the interdependence of the knobs, while TurBO's trust region iteration changes the size of the trust regions along each of the input axes in isolation. The proposed domain knowledge guided Bayesian Optimization approach performs better than the others and its median performance is able to reach the threshold beam position error to ensure overlap.

In Figure \ref{fig:bpe_comparison}, we report the results over the sets of trials for each algorithm for the Beam Intensity. Here, we observe that the proposed domain knowledge guided Bayesian Optimization approach leads to better results than the others. This superior performance is not merely a consequence of faster convergence on a single objective, but rather a direct result of the proposed approach's two-component strategy, which sequentially solves the problem's distinct challenges. The first component, the coordinate transformation, is responsible for the rapid minimization of the beam position error seen in Figure \ref{fig:bpe_comparison}. By aligning the optimizer's search axes with the problem's natural differential and common modes, the approach efficiently identifies the low-dimensional submanifold where the BPE is minimal. Crucially, the global intensity optimum is known to reside within this submanifold. This localization effectively reduces the dimensionality of the search space for the intensity peak, a phenomenon that manifests as a distinct two-phase behavior in the optimization trajectory. In the initial phase of the optimization (approximately the first 60 evaluations), the algorithm prioritizes finding this manifold, resulting in a rapid descent in the BPE while the measured intensity remains largely stationary. Once the search has converged onto this low-error region, the optimization enters a second phase (from approximately 60 to 150 evaluations), where the BPE is asymptotically stable and the algorithm's budget is redirected to exploring within this manifold, leading to a marked increase in beam intensity.

However, successfully localizing the search to this submanifold is only half the solution, as the intensity objective still presents a needle in a smaller haystack problem. A conventional optimizer, having found the broad region of low BPE, would tend to cease exploration and converge prematurely on a suboptimal point with negligible intensity. This is where the second component of the proposed domain knowledge guided Bayesian Optimization approach, reverse annealing, becomes indispensable. By systematically increasing the exploration hyperparameter, $\beta$, of the UCB acquisition function, the algorithm is forced to counteract its natural tendency to exploit the first good region it finds. This sustained exploratory pressure compels the optimizer to continue probing areas of high uncertainty within the low-error manifold, which is precisely where the undiscovered intensity peak is expected to lie. This forced exploration is directly responsible for the consistent increase in intensity observed during the second phase of optimization, enabling the proposed domain knowledge guided Bayesian Optimization approach to reliably escape local optima and locate the true global maximum.

The stagnation of TurBO algorithm's performance, even with its trust-region approach, can be attributed directly to the non-axis-aligned nature of the problem. As shown in Figure 3a, the diagonal canyon of the BPE objective is a fundamentally inefficient geometry for TurBO's axis-aligned rectangular trust regions, leading to wasted evaluations and a failure to make progress. The failure of the standard BO is a case of premature convergence. It is able to identify the basin of attraction corresponding to low BPE with additional samples. However, lacking the exploratory pressure of reverse annealing, standard BO focuses all subsequent evaluations on exploiting this large but suboptimal region, never discovering the sparse intensity peak. 

\begin{figure}[h!]
\centering
\includegraphics[width=\textwidth,trim={6.25cm 2.05cm 6.25cm 2cm},clip]{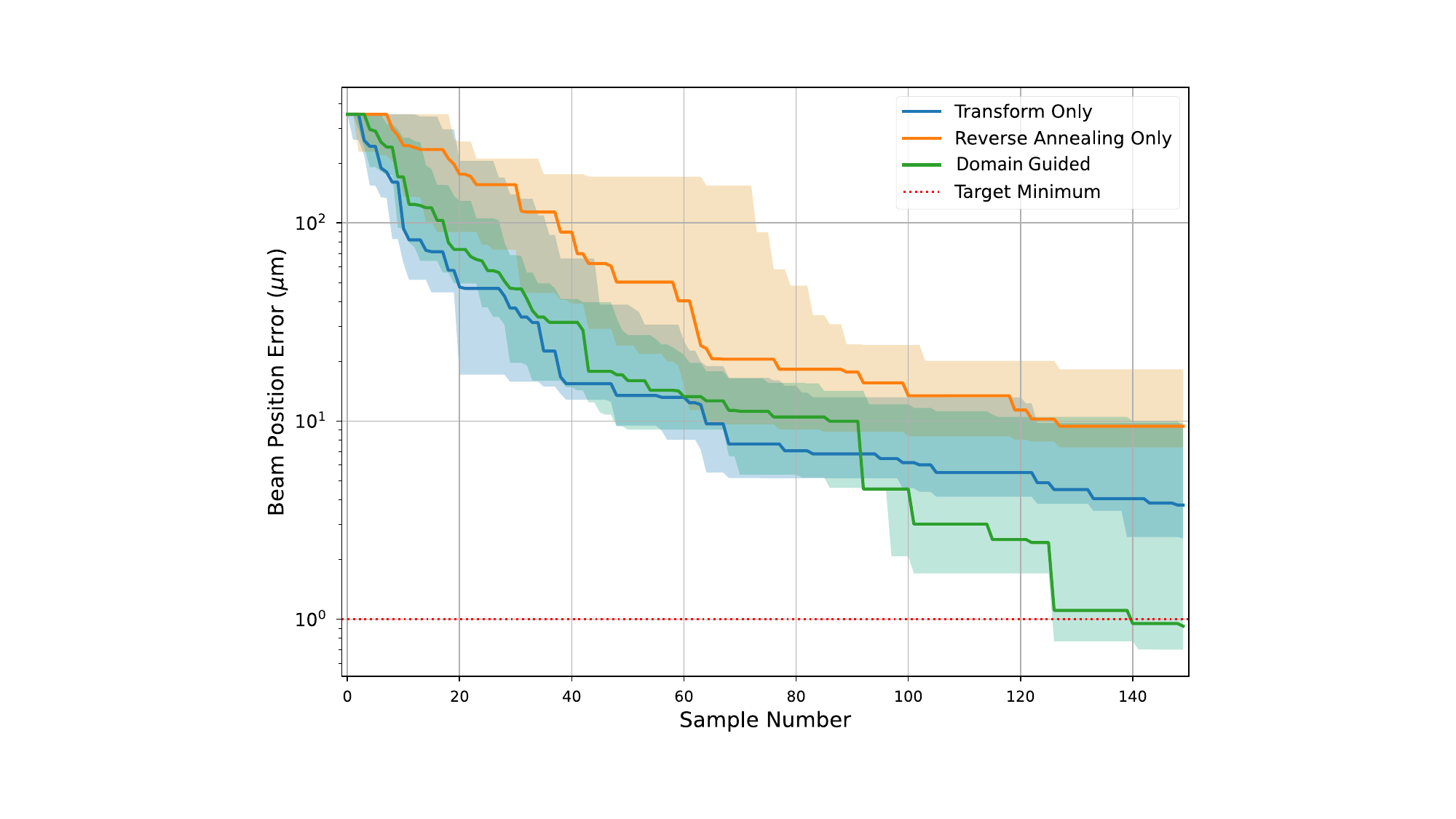}
\caption[]{Comparison between the performance of domain knowledge guided Bayesian Optimization, against versions with only the input co-ordinate transform and only the reverse annealing for the Beam Position Error. For each algorithm, we report the median of the running minimum with the solid line, over $25$ experiments. The shaded zones outline the $25^{th}$ and $75^{th}$ percentiles.}
\label{fig:bpe_ablation}
\end{figure}

\begin{figure}[h!]
\centering
\includegraphics[width=\textwidth,trim={6.25cm 2.05cm 6.25cm 2cm},clip]{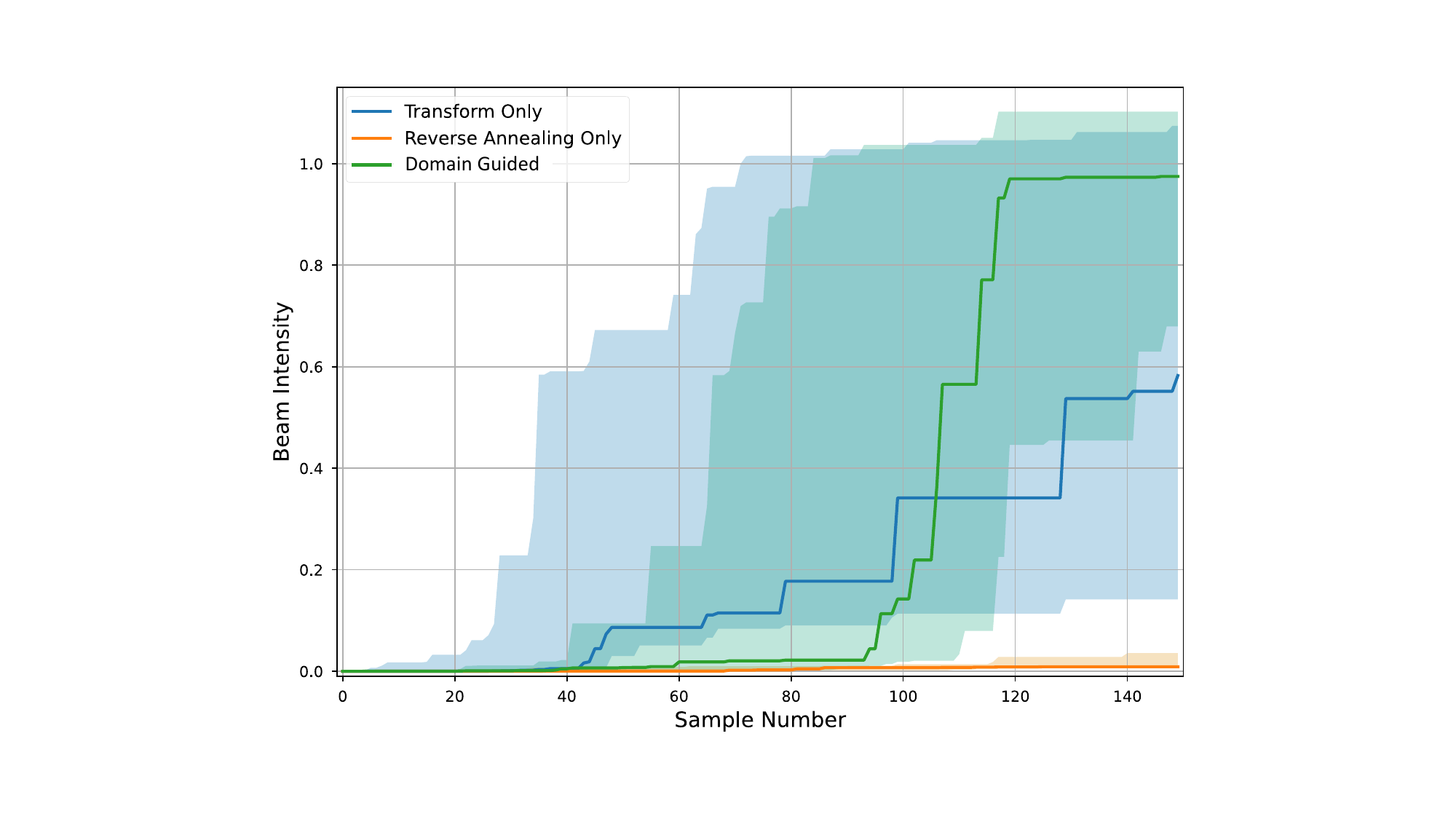}
\caption[]{Comparison between the performance of domain knowledge guided Bayesian Optimization, against versions with only the input co-ordinate transform and only the reverse annealing for the Beam Intensity. For each algorithm, we report the median of the running maximum with the solid line, over $25$ experiments. The shaded zones outline the $25^{th}$ and $75^{th}$ percentiles.}
\label{fig:intensity_ablation}
\end{figure}

The domain knowledge guided Bayesian Optimization approach consists of two components: the input co-ordinate transform to align the input axes with the active subspaces, and for this HXRSND problem, the reverse annealing to increase the tendency for exploration, especially in the later stages of optimization. These exploit the complex but symbiotic character of the beam position error and intensity optima. While the intensity optima represents a needle in a haystack problem due to its sparsity, achieving the beam position error minimum may help reduce the size of the proverbial haystack as the intensity optimum lies on the manifold of the beam position error minimum. Similarly, achieving the maximum of the intensity leads to the optimization of the beam position error as well. To understand the individual contributions of the two components (the coordinate transformation and the reverse annealing schedule), we conducted an ablation study. We evaluated the performance of two additional variants: a Transform Only variant using a fixed exploration parameter, and an Annealing Only variant operating in the un-transformed coordinate space. The results are compared against the full domain knowledge guided Bayesian Optimization implementation in Figures \ref{fig:bpe_ablation} and \ref{fig:intensity_ablation}. The results of the ablation study demonstrate the necessity of both components in the procedure. The Transform Only variant rapidly minimizes the BPE but consistently fails to locate the intensity peak and converges prematurely to a suboptimal solution. This confirms that correctly structuring the search space, while helpful, is still insufficient on its own to solve the entire needle in a haystack problem. Conversely, the Annealing Only variant is highly inefficient, struggling to even find the low-error submanifold within the allotted sample budget due to the non-axis-aligned problem structure. Only the full domain knowledge guided Bayesian Optimization implementation which leverages the transformation to efficiently find the target manifold and the reverse annealing to explore within it reliably converges to the true global optimum. This confirms that our two-part strategy successfully and sequentially addresses the dual challenges of the optimization landscape.

\section{Conclusion \& Broader Impact}
In this work, we addressed a fundamental weakness of Bayesian Optimization and its advanced variants: their frequent failure in high-dimensional, needle in a haystack problems where tightly coupled parameters create complex, non axis-aligned objective landscapes. While BO excels at learning from limited data, forcing it to discover these underlying correlations from scratch is inefficient, unreliable and often leads to premature convergence in suboptimal regions. 

To overcome this, we outlined and validated a generalizable framework, domain knowledge guided Bayesian Optimization. \textit{The core principle is not to build a more complex surrogate model or acquisition function, but to simplify the problem itself}. We accomplish this via a two-pronged strategy: (1) a domain knowledge derived coordinate transformation that decouples parameter interdependencies and aligns the search directions with the problem's natural active subspaces, and (2) a reverse annealing schedule that ensures robust exploration of the now simplified landscape. We demonstrated the power of this paradigm through a rigorous case study on a challenging 12-dimensional optical alignment system. On this task, where standard BO, Trust Region BO (TuRBO), and multi-objective approaches (MOBO) consistently led to unsatisfactory results, our method converged reliably and efficiently to the global optimum.

The applicability of this framework extends beyond X-ray optics, as such challenges are ubiquitous across complex scientific instruments. For example, achieving a stable beam orbit in a particle accelerator requires co-optimizing hundreds of steering magnets whose effects are coupled through the machine's optics lattice. Here, our framework provides a direct path to use the known response matrix as a transformation for more efficient optimization. Similarly, this methodology is directly applicable to adaptive optics, where the coupled influence functions of a deformable mirror's actuators can inform a transformation to more rapidly correct wavefront aberrations. In such cases, by using domain knowledge to transform the problem's geometry rather than forcing an algorithm to learn complex correlations from scratch, this methodology promises a advantage: superior sample efficiency, greater robustness against premature convergence, and the consistent attainment of higher-quality optima. This approach of encoding domain knowledge as an explicit inductive bias, not in the model architecture, but in the definition of the search space itself, represents a practical step towards creating more intelligent and autonomous scientific systems.

\begin{funding}
Use of the Linac Coherent Light Source (LCLS), SLAC National Accelerator Laboratory, is supported by the U.S. Department of Energy, Office of Science, Office of Basic Energy Sciences under Contract No. DE-AC02-76SF00515.
\end{funding}

{The authors declare no conflicts of interest.
}

\ConflictsOfInterest{The authors declare no conflicts of interest.
}

\DataAvailability{The data and the code underlying the results presented in this paper are not publicly available at this time but can be obtained from the authors upon request.}

\appendix
\section{Appendix A: The domain knowledge guided Bayesian Optimization Approach - a general recipe}
The core philosophy of this domain knowledge guided Bayesian Optimization approach is to leverage existing domain physics knowledge to simplify the geometry of the search problem \textit{before} the optimization. Whereas standard optimization algorithms must learn complex correlations, skewed objective landscapes, active subspaces, etc from scratch, the proposed workflow transforms the problem into a ``native" coordinate system where the parameters are more decoupled and thus, the path to the optimum is more direct. This leads to superior sample efficiency and robustness. Applying this methodology to a new problem can be formulated as a five-step recipe:

\textit{Step 1: Identify Control and Performance Spaces}
First, explicitly define the native control space representing the set of $n$ parameters that can be directly actuated (for instance, motor positions, voltages, magnet currents). Then, define the objective or performance metric $y = f(x)$ that the optimization seeks to maximize or minimize (for instance, beam intensity, aerodynamic drag, plasma stability, etc).

\textit{Step 2: Identify the Underlying Physical Model or Symmetries}
The ``domain knowledge guided" component comes from identifying a known, simplified relationship between the control parameters and the system's physical state. This knowledge does not need to be a perfect model of the objective function $f(x)$; it only needs to describe the underlying correlations in the control space. This can take the form of Linear Response Models, analytical models, etc.

\textit{Step 3: Define a domain knowledge informed Coordinate Space}
Using the insight from Step 2, define a new, physically meaningful coordinate space. The goal is for the axes of this new space to align with the active subspaces, or the natural symmetries of the problem. For instance, instead of controlling four individual motor positions ($x = [m_1, m_2, m_3, m_4]$), one might define a new input feature based on their collective physical effect ($x' = $[average height, vertical tilt, horizontal tilt, total torsion]). This new set of coordinates $x'$ is related to the native controls $x$ by a transformation $T$: $x' = T(x)$.

\textit{Step 4: Formulate and Invert the Transformation}
The transformation T can range from a simple linear map (a matrix multiplication, $x' = A x$) to a more complex non-linear function. The Bayesian optimization algorithm will now search for the optimum in the transformed space, proposing the next point to evaluate as $x'_{next}$. To perform the physical experiment, this point must be translated back into the native control space. This requires computing the inverse transformation: $x_{next} = T^{-1}(x'_{next}$. This inverse mapping must be well-defined and computationally efficient. For a linear transformation, this is typically a straightforward matrix inversion or solving a system of linear equations.

\textit{Step 5: Execute Bayesian Optimization in the Transformed Space}
Finally, execute the Bayesian optimization loop using the transformed coordinates. The Gaussian Process model is built on observations of $(x', y)$ pairs. The acquisition function then proposes the next candidate point $x'_{next}$ in the simplified space, which is then inverted back to $x_{next}$ for evaluation on the real system. As the optimization algorithm is now exploring a space where the objective function's features are better aligned with the axes, it can locate the optimum far more efficiently, consistently and robustly.

\section{Appendix B: Simulation Details \& Validation}
\begin{figure}
        \centering
        \includegraphics[width=\textwidth]{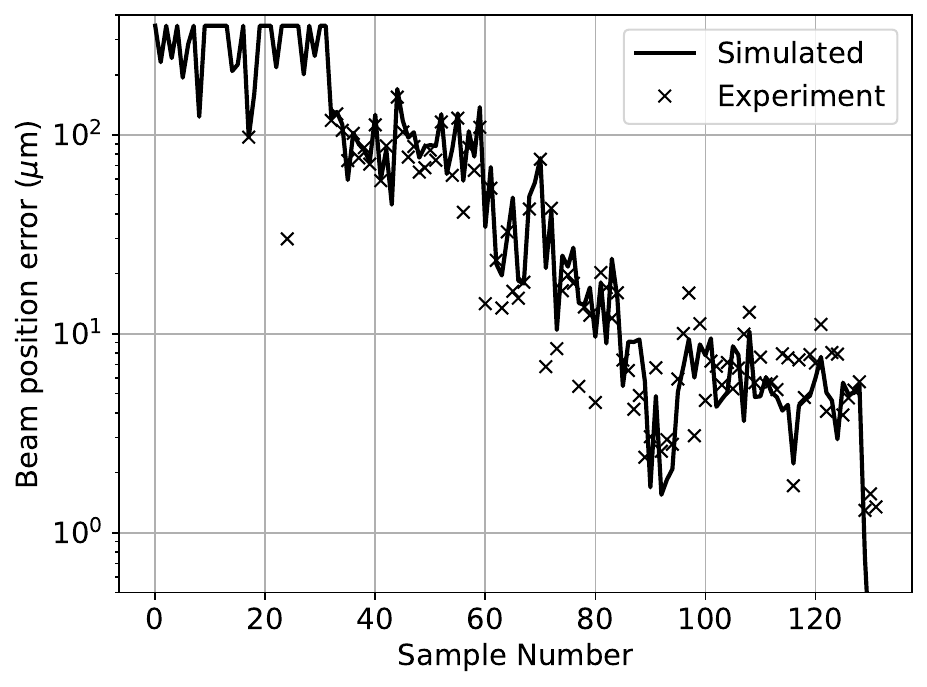}
        \caption{Comparison of the predictions of the Beam Position Error from the wave-optical simulations code used in this investigation against experimental results on the HXRSND (delay branch only). Between each sample various degrees of freedom were adjusted both in experiment and also as inputs to the simulator.}
        \label{fig:simulation_validation}
\end{figure}

For this study, we performed wave-optical simulations of the system in preparation of the limited availability of XFEL beamtime. The simulations are based on decoupled horizontal and vertical (2$\times$1D) wavefront propagation using in-house software\cite{lcls_beamline_toolbox}. These simulations model the input beam as fully spatially coherent (a reasonable assumption for XFEL beams) and monochromatic (taking credit for a monochromator upstream of the split and delay system). The various motion degrees of freedom shown in Figure~\ref{fig:fig1} are all reproduced in the simulation, and care was taken to ensure that the simulation sufficiently captures the dynamics of the operation of the actual HXRSND (see Figure~\ref{fig:simulation_validation}). Since the CC branch is intrinsically much more stable than the delay branch, we focus the simulation efforts on maintaining the stability of the delay branch. In this study, the system was configured for operation at 9.5 keV with zero relative delay between the branches. To judge spatial overlap, we simulate the position of the beam directly at the interaction point as if it were measured using YAG fluorescence. Even though this represents an invasive measurement, the results of the simulation study also apply to non-invasive measurements that are under development, assuming they provide equivalent information.


\begin{figure}
        \centering
        \includegraphics[width=\textwidth]{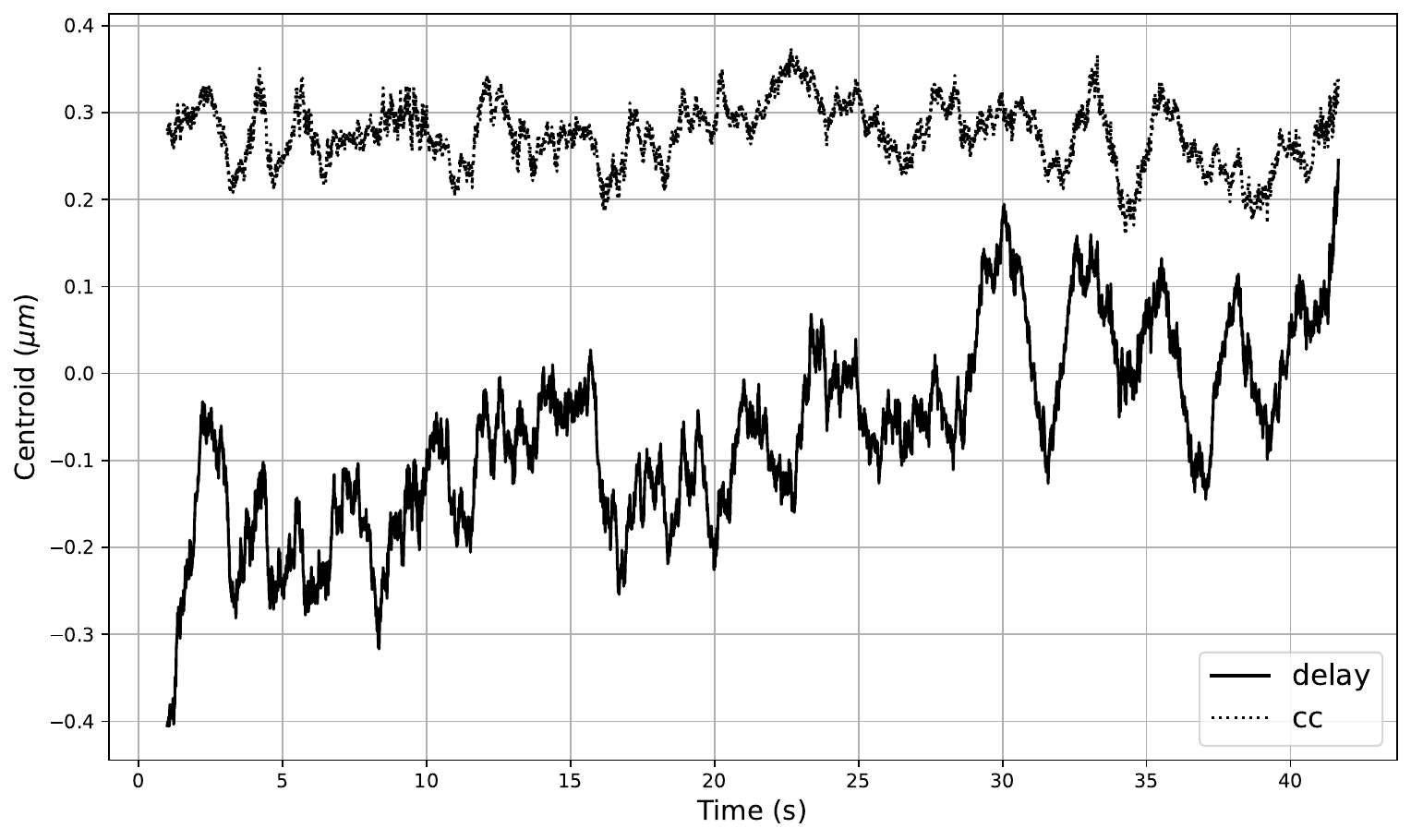}
        \caption{Visualization of the drift for the horizontal Beam Position Error in the Channel Cut (labeled as cc) and the Delay (labeled as delay) branches observed in an experiment using the HXRSND, without adjustment of any of the degrees of freedom. The rate of drift is observed to be $\approx$300~nm in 1 minute, and the standard deviation after subtracting the linear drift is 108~nm rms.}
        \label{fig:experimentaldriftrate}
\end{figure}

\bibliography{iucr} 

\end{document}